\documentclass{article}
\usepackage{spconf,amsmath,graphicx}

\usepackage{xcolor}
\usepackage{multirow}
\usepackage{subfigure}
\usepackage{graphicx}
\usepackage{mathtools}
\usepackage{eqparbox}
\usepackage{amssymb}

\usepackage[linesnumbered,ruled,vlined]{algorithm2e}
\usepackage{algpseudocode}
\usepackage{amsmath}
\usepackage{array}

\SetKwComment{Comment}{ $\triangleright$ }{}

\title{{LANCE: Efficient Low-Precision Quantized Winograd Convolution \\for Neural Networks based on Graphics Processing Units}}
\name{Guangli Li$^{\S\dag}$, Lei Liu$^{\S}$, Xueying Wang$^{\S\dag}$, Xiu Ma$^{\S\ddag}$, and Xiaobing Feng$^{\S\dag}$ \thanks{Special thanks to Dr.~Chong Li for his constructive comments and suggestions. This work is supported by the National Key R\&D Program of China under Grant No.2017YFB1003103, and the Science Fund for Creative Research Groups of the National Natural Science Foundation of China under Grant No.61521092. \textit{(Corresponding author: Lei Liu.)}}}
\address{\normalsize
$^{\S}$SKL of Computer Architecture, Institute of Computing Technology, Chinese Academy of Sciences, China\\ \normalsize
$^{\dag}$School of Computer Science and Technology, University of Chinese Academy of Sciences, China\\ \normalsize
$^{\ddag}$College of Computer Science and Technology, Jilin University, China
\\ \normalsize
\{liguangli, liulei, wangxueying, maxiu01, fxb\}@ict.ac.cn
}

\usepackage{fancyhdr}

\begin{document}
\topmargin=0mm

\maketitle

\thispagestyle{fancy}
\fancyhead{}
\rhead{}
\lfoot{\footnotesize \copyright~2020 IEEE.  Personal use of this material is permitted.  Permission from IEEE must be obtained for all other uses, in any current or future media, including reprinting/republishing this material for advertising or promotional purposes, creating new collective works, for resale or redistribution to servers or lists, or reuse of any copyrighted component of this work in other works.}
\cfoot{}
\rfoot{}

\begin{abstract}
Accelerating deep convolutional neural networks has become an active topic and sparked an interest in academia and industry.
In this paper, we propose an efficient low-precision quantized Winograd convolution algorithm, called \textsc{Lance}, which combines the advantages of fast convolution and quantization techniques.
By embedding linear quantization operations into the Winograd-domain, the fast convolution can be performed efficiently under low-precision computation on graphics processing units.
We test neural network models with \textsc{Lance} on representative image classification datasets, including SVHN, CIFAR, and ImageNet.
The experimental results show that our 8-bit quantized Winograd convolution improves the performance by up to 2.40$\times$ over the full-precision convolution with trivial accuracy loss.

\end{abstract}
\vspace{-0.2cm}
\section{Introduction}
\vspace{-0.1cm}
\label{sec:intro}

Deep convolutional neural networks (DCNNs) have adequately advanced diverse intelligent applications
such as image classification \cite{krizhevsky2012imagenet} and object detection \cite{ren2015faster}.
While sophisticated neural networks are effective at continuously improving the accuracy of intelligent tasks,
the computational complexity, as well as the storage requirement, is also drastically increased,
leading to an obstacle for their applicability.
Real-time applications, such as video surveillance, have a strict constraint of processing time, and embedded applications, such as virtual reality, a limitation of memory usage.
As such, the research of accelerating and compressing neural networks becomes an inevitable trend.
On the one hand, fast convolution algorithms for neural networks, such as FFT (fast Fourier transform) \cite{mathieu2013fast} and Winograd's minimal filtering algorithm  \cite{lavin2016fast}, are proposed, which reduce the number of computations in convolution by exploiting the correspondence between the convolution and the scalar multiplication.
On the other hand, there are several approaches focus on compressing neural networks by using quantization techniques \cite{han2015deep}\cite{guo2018survey}, which represents original floating-point values by low-precision (e.g., 8-bit integer) codes.
The quantized values can be executed under low-precision computation that has the potential of speedup~\cite{jacob2018quantization}.
Unfortunately, these two kinds of techniques cannot be directly combined because the data transformation of fast convolution algorithms disturbs the quantized values, which eliminates the gain of low-precision quantization.
In this paper, we address the above problem by proposing \textsc{Lance} (\textbf{L}ow-precision qu\textbf{A}ntized \textsc{w}i\textbf{N}ograd \textbf{C}onvolution for neural n\textbf{E}tworks), which applies quantization methods into the Winograd-domain.

\begin{figure*}[htbp]
\centering
\begin{tabular}{c|c}
\subfigure[Winograd convolution in quantized neural networks.]{
\label{ME-1}
\begin{minipage}[t]{0.43\textwidth}
\centering
\includegraphics[height=4.8cm]{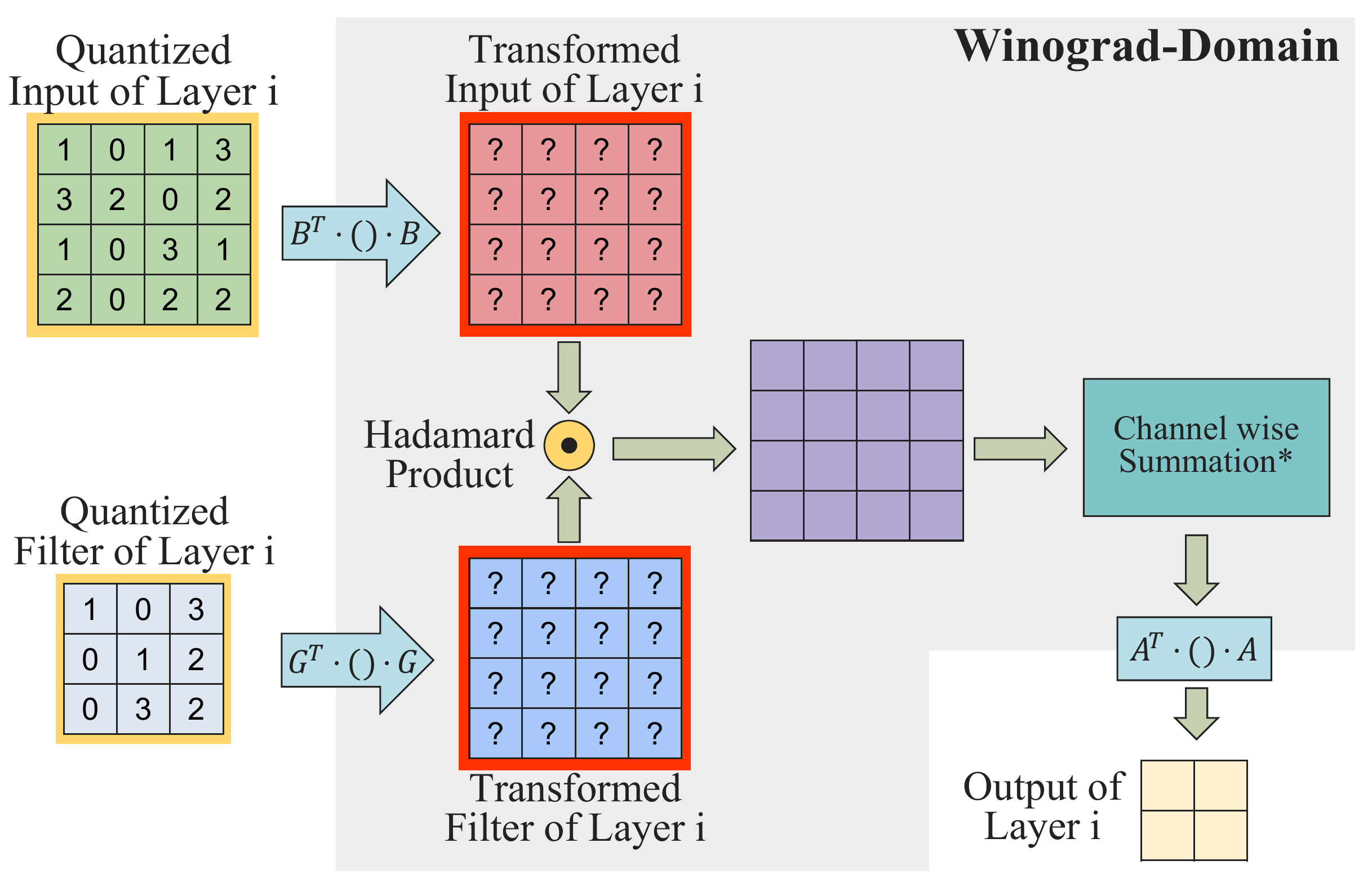}
\end{minipage}
}
&
\subfigure[Our proposed quantized Winograd convolution.]{
\label{ME-2}
\begin{minipage}[t]{0.51\textwidth}
\centering
\includegraphics[height=4.8cm]{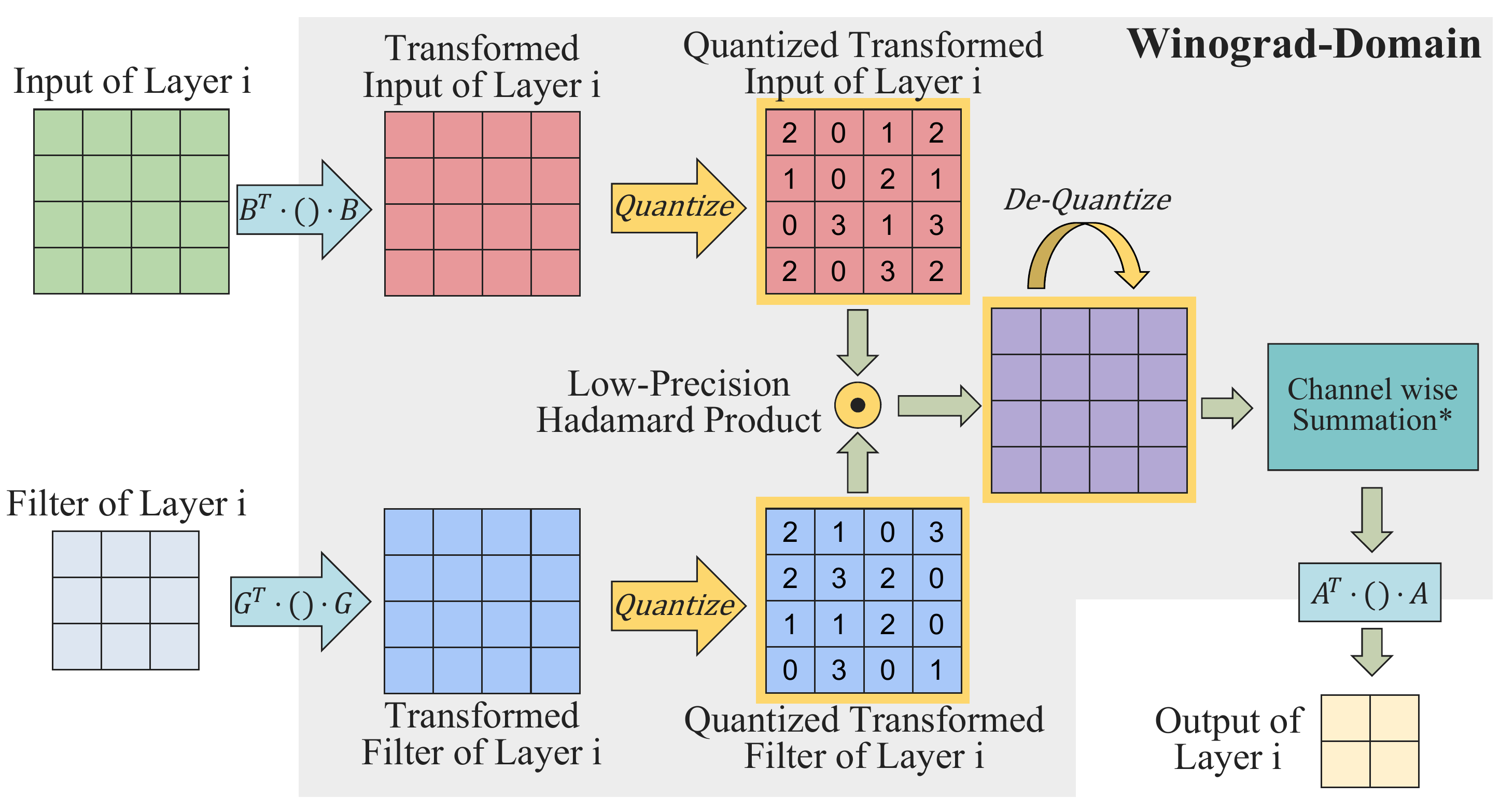}
\end{minipage}
}
\end{tabular}
\caption{Combining Winograd convolution with quantization: brute-force approach vs our approach.}
\end{figure*}

\vspace{-0.1cm}
\section{Preliminary and Formalization}
\vspace{-0.1cm}
\subsection{Convolution Computation}

Convolution is a crucial operation in deep convolutional neural networks,
which extracts features from input images by applying several filters and obtains the results into output images (also known as feature maps).
We denote the input images as $X$, the convolution filters as $W$, and the output images as $Y$.
A typical convolution layer of deep neural networks (stride and padding are omitted) can be presented by:

\begin{equation}
\label{e1}
\centering
\small
{
    Y_{n,k,i,j} = \sum^{C}_{c=1} \sum^{R}_{r=1} \sum^{S}_{s=1} X_{n,c,i+r,j+s} \times W_{k,c,r,s}
}
\end{equation}
$X_{n,c,i+r,j+s}$ is the element on $(i+r)_{th}$ row and $(j+s)_{th}$ column in the $c_{th}$ input channel of $n_{th}$ input image,
$W_{k,c,r,s}$ is the element on $r_{th}$ row and $s_{th}$ column in the $c_{th}$ channel of $k_{th}$ filter,
and $Y_{n,k,i,j}$ is the element on $i_{th}$ row and $j_{th}$ column in the $k_{th}$ output channel of $n_{th}$ output image.
For an entire image/filter pair, the equation can be expressed as:
\begin{equation}
\label{e2}
\centering
\small
{
    Y_{n,k} = \sum^{C}_{c=1} X_{n,c} * W_{k,c}
}
\end{equation}
where $*$ represents the 2D correlation (refer to \cite{lavin2016fast}).

\subsection{Low-precision Quantization}

Low-precision quantization techniques represent the values of original data by low-precision quantized codes~\cite{cheng2018recent}.
In general, the process of the convolution with quantization has three steps:
Firstly, converting the values of images and filters to quantized values by using a quantization function $Q$;
Then, performing the low-precision computation with quantized values;
Finally, converting the values of quantized results to output feature maps using a de-quantization function $Q'$.
Thus, the quantized convolution can be formalized as:
\vspace{-0.1cm}
\begin{equation}
\label{e3}
\centering
\small
{
    Y_{n,k} = \sum^{C}_{c=1} Q'\left(Q\left(X_{n,c}\right) \circledast Q\left(W_{k,c}\right)\right)
}
\end{equation}
where $\circledast$ represents the quantized 2D correlation which can be calculated under low-precision computation.

\subsection{Winograd Convolution}

Similar to \cite{lavin2016fast}, we use $F(m \times n, r \times s)$ to denote the Winograd convolution of an $m \times n$ output with an $r \times s$ filter.
$F(m \times n, r \times s)$ requires $(m + r - 1)(n + s - 1)$ multiplications~\cite{winograd1980arithmetic}, which equals to the number of input elements, whereas the standard convolution requires $(m \times n \times r \times s)$ multiplications.
For the 2D Winograd convolution $F(m \times m, r \times r)$, the basic block of output is an $m \times m$ patch and input an $(m+2) \times (m+2)$ patch which extracted from input images.
An input image is divided into several patches (with stride and padding if necessary) and the corresponding output patches are merged in an output feature map.
Let the input patch is $d$, the filter is $g$, and the output patch is $S$, the Winograd algorithm can be written as:
\vspace{-0.1cm}
\begin{equation}
\label{e-wino}
\centering
\small
{
    S = A^\mathrm{T}\left[ \left[GgG^\mathrm{T}\right] \odot \left[B^\mathrm{T} dB\right]\right]A
}
\end{equation}
where $\odot$ represents the Hadamard product.
$A$, $B$, and $C$ are the transformation matrices.
$GgG^\mathrm{T}$ and $B^\mathrm{T} dB$ are the transformed filter and transformed input patch in the Winograd-domain, respectively.
By using the matrices $A$ and $A^\mathrm{T}$, $S$ is obtained.
The detailed algorithm can be seen in \cite{lavin2016fast}.

\vspace{-0.1cm}
\section{Proposed Approach}
\vspace{-0.1cm}
\label{sec:lance}
\subsection{A Motivating Example}
\label{sec:example}

Let $F(2 \times 2, 3 \times 3)$ as an example (i.e. Winograd convolution with a $4 \times 4$ input $d$, a $3 \times 3$ filter $g$, and a $2 \times 2$ output $S$), which is widely used in modern neural networks.
Figure \ref{ME-1} shows the brute-force approach of combining the Winograd convolution with quantization techniques.
In quantized neural networks, the input and filters of a convolution layer are quantized to low-precision codes, as the green matrix and gray matrix.
However, the transformation operations of the Winograd convolution disturb the quantized codes, which means that the values of transformed matrices are indeterminate or cannot be represented by existing quantized codes in the Winograd-domain (the red matrix and blue matrix).
For demonstration purposes, we use a naive 2-bit linear quantization method (quantized codes$\in$\{$0,1,2,3$\}).
Let the original full-precision input is $d$.
By using the function $Q$, the quantized input:
\vspace{-0.1cm}
\begin{equation}
{
\centering
\footnotesize
\setlength{\arraycolsep}{3pt}
\hat{d}=Q(d)=Q\left(
\begin{bmatrix*}[r]
0 & 1  & 2 & 3  \\
4 & 5  & 6 & 7  \\
8 & 9  & 10  & 11  \\
12 & 13 & 14  & 15
\end{bmatrix*}
\right)
=\begin{bmatrix*}[r]
0 & 1 & 1 & 1  \\
1 & 1 & 2 & 2  \\
2 & 2 & 2 & 3  \\
3 & 3 & 3 & 3
\end{bmatrix*}
}
\end{equation}
The transformed input matrix:
\vspace{-0.1cm}
\begin{equation}
{
\centering
\footnotesize
\setlength{\arraycolsep}{2pt}
B^\mathrm{T}\hat{d}B=
\begin{bmatrix*}[r]
1 & 0  & -1 & 0  \\
0 & 1  & 1  & 0  \\
0 & -1 & 1  & 0  \\
0 & 1  & 0  & -1
\end{bmatrix*}\hat{d}B
=\begin{bmatrix*}[r]
-1 & -2 & 0 & 1  \\
-1 & 7 & 1 & -2  \\
1 & 1 & -1 & 0  \\
-1 & -3 & 1 & -1
\end{bmatrix*}
\label{eqn-trans}
}
\end{equation}
As can be seen, the values of the transformed matrix cannot be represented by the above-mentioned 2-bit low-precision codes.
What's more, if we use full-precision data types to present the transformed result, however, the Hadamard product cannot use the potential of low-precision computation.

\noindent \textbf{Remarks.}
The example of brute-force approaches shows that the values of quantized input images and filters are disturbed by the Winograd transformations, which cannot be directly used in the Winograd-domain.
What we need is a method that can combine the Winograd algorithm with quantization techniques to achieve more with their advantages.

\subsection{Low-Precision Quantized Winograd Convolution}
As shown in Figure \ref{ME-2}, we propose \textsc{Lance}, which applies quantization techniques into the Winograd-domain, to explore the potential of low-precision computation.
In our algorithm, we use a uniform linear quantization function $Q$ to quantize a full-precision value $x$ to a low-precision code~$\hat{x}$:
\vspace{-0.1cm}
\begin{equation}\label{q-func}
\centering
\small
{
\hat{x} = Q(x)= round\left(\frac{\left(x - \min(T)\right)\times \left(2^b-1\right)}{\max(T)- \min(T)}\right)
}
\end{equation}
where $b$ indicates the bits of low-precision data type, $T$ is the matrix that the value $x$ belongs to, such as input patch matrix $d$ and filter matrix $g$.
The quantized code $\hat{x}$ can be recovered to original full-precision by a de-quantization function $Q'$:
\vspace{-0.1cm}
\begin{equation}\label{q-defunc}
\centering
\small
{
\tilde{x} = Q'(\hat{x})= \hat{x}\times \left( \frac{\max\left(T\right)- \min(T)}{2^b-1}\right)+\min(T)
}
\end{equation}
Overall, the approach of the quantized Winograd convolution can be formalized as:
\vspace{-0.1cm}
\begin{equation}
\label{e-quantized-wino}
\centering
\small
{
    S = A^\mathrm{T}Q'\left(\left[ \left[Q\left(GgG^\mathrm{T}\right)\right] \odot \left[Q\left(B^\mathrm{T}dB\right)\right]\right]\right)A
}
\end{equation}
As such, the Hadamard product with quantized operands, $Q\left(GgG^\mathrm{T}\right)$ and $Q\left(B^\mathrm{T}dB\right)$, can be calculated under low-precision computing, which addresses the problems in the brute-force approach.

\vspace{-0.2cm}
\begin{algorithm}[!ht]
\small
    \SetCommentSty{itshape}
    \caption{\textsc{Lance}}
    \label{algo1}
    \KwIn{$X$ (input image), $W$ (filters)}
    \KwOut{$Y$ (output feature map)}
    $K \leftarrow$ \#\textit{Filters} of $W$ \label{l-1}\;
    $P \leftarrow$ \#\textit{Patches} of $X$ per channel \label{l-2}\;
    $C \leftarrow$ \#\textit{Channels} of $X$ and $W$ \label{l-3}\;
    $D \leftarrow$ Patches divided from $X$ with padding \label{l-4}\;
    \For{$k\leftarrow 1$ \KwTo $K$\label{l-5}}{
    \For{$p\leftarrow 1$ \KwTo $P$\label{l-6}}{
        $m \leftarrow O$\;
    \For{$c\leftarrow 1$ \KwTo $C$\label{l-7}}{
        $g \leftarrow W_{k,c}$ ($k_{th}$ filter at $c_{th}$ channel) \label{l-8}\;
        $d \leftarrow D_{p,c}$ ($p_{th}$ patch at $c_{th}$ channel) \label{l-9}\;
        $\hat{u} \leftarrow Q(G g G^\mathrm{T})$ \label{l-10} \Comment*[r]{Quantize values}
        $\hat{v} \leftarrow Q(B^\mathrm{T} d B)$ \label{l-11} \Comment*[r]{Quantize values}
        $\hat{m} \leftarrow \hat{u} \odot \hat{v}$ \label{l-12} \Comment*[r]{Low-precision computation}
        $\tilde{m} \leftarrow Q'(\hat{m})$ \label{l-13} \Comment*[r]{De-quantize the result}
        $m \leftarrow m + \tilde{m}$ \label{l-14} \Comment*[r]{Channel-wise
summation}
    }
        $S \leftarrow A^\mathrm{T}mA$ \label{l-15}\;
        $Y_{p,k} \leftarrow S$ \label{l-16}\;
    }
    }
    \Return $Y$ \label{l-17}\;
\end{algorithm}
\vspace{-0.2cm}

Algorithm \ref{algo1} describes our quantized Winograd algorithm for a convolution layer.
The input of \textsc{Lance} are an image $X$ and filters $W$, and the output feature maps, which initialized by zero.
Here, we assume the batch size of the neural network $N=1$.
The approach for larger batch size, $N>1$, remains the same.
The transformation matrices, $G$, $B$, and $A$, are generated based on Chinese
Remainder Theorem with given sizes of the input and filters.
Firstly, the number of filters, patches, and channels are obtained from input data (Lines \ref{l-1}-\ref{l-3}).
The input image $X$ is divided and recorded in $D$ (Line \ref{l-4}).
Each input patch $d$ and filter $g$ are transformed by using $G$ and $B$ (Lines \ref{l-8}-\ref{l-9}).
The transformed values are quantized to low-precision codes (Lines \ref{l-10}-\ref{l-11}).
The Hadamard product of quantized transformed data, $\hat{u}$ and $\hat{v}$, is calculated by low-precision computing (Line \ref{l-12}).
The output patch $S$ is transformed from the sum of Hadamard product results by using $A$ (Lines \ref{l-13}-\ref{l-15}).
Finally, the output feature map $Y$ is obtained, which is merged by output patches~(Lines \ref{l-16}-\ref{l-17}).

\vspace{-0.3cm}
\subsection{Implementation on Graphics Processing Units}

We describe an efficient implementation of \textsc{Lance} on graphics processing units (GPUs), which are commonly used to accelerate intelligent applications in recent years.

\noindent \textbf{Data Layout}
We use NHWC, a common format in deep learning applications, as the data layout of \textsc{Lance}, where N denotes the batch size, H denotes the height of input images, W denotes the width of input images, and C denotes the channel of input images, respectively.
The same pixels in different channels are stored contiguously, which can be processed simultaneously, and the parallelizability of these pixels is not influenced by the bit-widths of data types.
Therefore, the NHWC format is more suitable for low-precision data types.

\noindent \textbf{Data Transformation and Hadamard Product}
For transformation of data, each thread calculates the transformed data of a patch with single channel and the quantization or de-quantization of transformed patches can be parallelized.
The Hadamard product is computed by using low-precision general matrix multiplication (GEMM) \cite{lavin2016fast}, which can be performed by effective specific instructions on GPUs.
We implement the low-precision GEMM by leveraging the WMMA~\cite{WP-turing} APIs of CUDA C++.

\vspace{-0.1cm}
\subsection{Embedding into Convolutional Neural Networks}
\vspace{-0.1cm}

\vspace{-0.3cm}
\begin{figure}[!ht]
  \centering
  \includegraphics[width=0.45\textwidth]{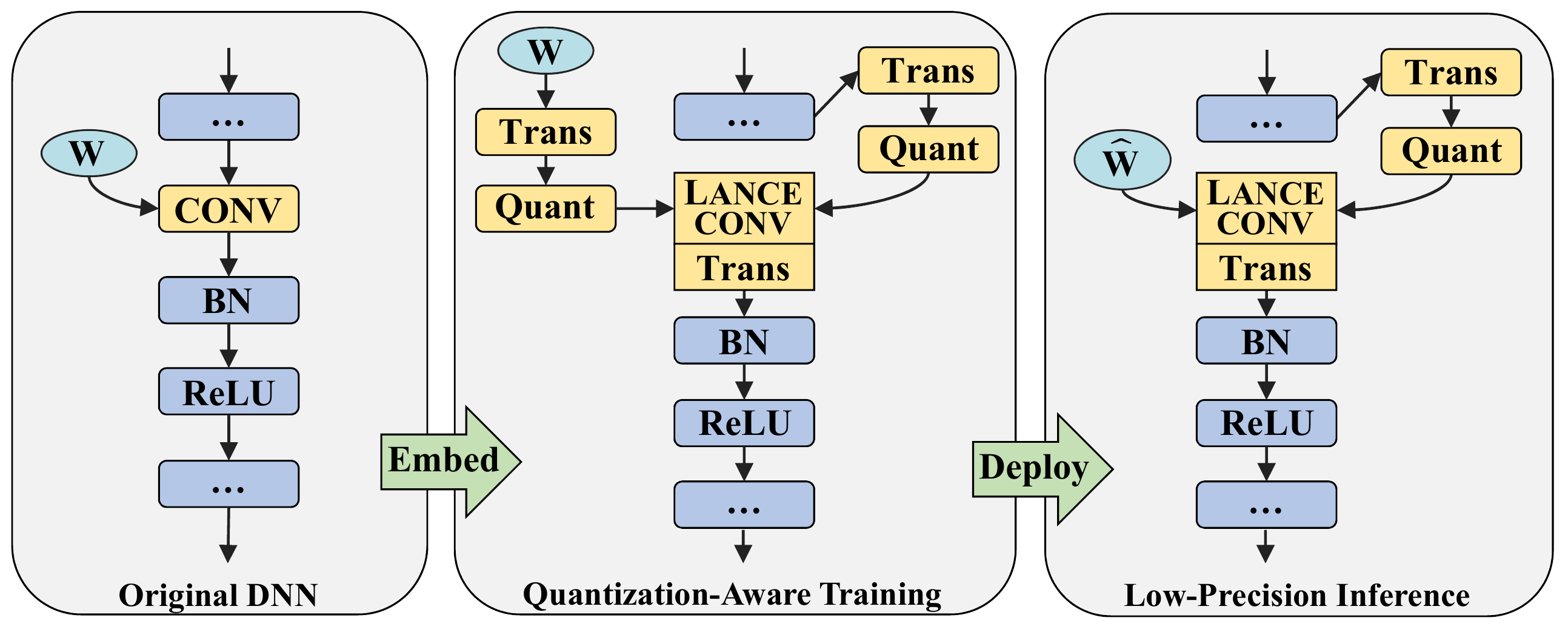}
  \caption{Applying \textsc{Lance} in neural networks.}
  \label{embed}
\end{figure}
\vspace{-0.2cm}

\noindent Figure \ref{embed} depicts the approach of applying our \textsc{Lance} in neural networks.
The convolution is replaced with our low-precision Winograd convolution and corresponding transformation and quantization operations are inserted.
We train the modified model by simulated quantization~\cite{jacob2018quantization}.
For inference, the transformation and quantization operations of weights in \textsc{Lance} are computed off-line just once, which can decrease the run-time overhead.

\vspace{-0.1cm}
\section{Experiments}
\vspace{-0.1cm}

We evaluate the performance of \textsc{Lance} on representative image classification datasets with different scales, including SVHN~\cite{Netzer2011Reading}, CIFAR~\cite{cifar-10}, and ImageNet 2012~\cite{Russakovsky2015ImageNet}.
The main traditional convolution layers of deep neural networks are replaced with our quantized Winograd convolution layers
and the hyper-parameters of training are the same for a specific network architecture.
The inference experiments are performed on a latest NVIDIA GPU, Tesla T4.

\setcounter{table}{0}
\vspace{-0.2cm}
\begin{table}[!ht]
\centering
\scriptsize
\begin{tabular}{|c|c|l|c|c||c|c|}
\cline{1-2} \cline{4-7}
\multicolumn{2}{|c|}{\textbf{SVHN}} &  & \multicolumn{4}{c|}{\textbf{CIFAR-10}} \\ \cline{1-2} \cline{4-7}
\multicolumn{2}{|c|}{\textbf{ConvNet-S}} &  & \multicolumn{2}{c||}{\textbf{ConvNet}} & \multicolumn{2}{c|}{\textbf{VGG-nagadomi}} \\ \cline{1-2} \cline{4-7}
\textbf{W-I} & \textbf{ACC} &  & \textbf{W-I} & \textbf{ACC} & \textbf{W-I} & \textbf{ACC} \\ \cline{1-2} \cline{4-7}
\textbf{32-32} & 0.9572 &  & \textbf{32-32} & 0.8912 & \textbf{32-32} & 0.9042 \\ \cline{1-2} \cline{4-7}
\textbf{8-8} & $\uparrow$ \textbf{0.11\%} &  & \textbf{8-8} & $\uparrow$ \textbf{1.04\%} & \textbf{8-8} & $\downarrow$ \textbf{0.08\%} \\ \cline{1-2} \cline{4-7}
\textbf{4-4} & $\uparrow$ 0.30\% &  & \textbf{4-4} & $\downarrow$ 2.43\% & \textbf{4-4} & $\downarrow$ 3.48\% \\ \cline{1-2} \cline{4-7}
\end{tabular}
\caption{The results on SVHN and CIFAR-10: W (weight bits), I (input bits), ACC (accuracy).}
\label{result-svhn}
\end{table}
\vspace{-0.2cm}

Table \ref{result-svhn} illustrates the result of ConvNet-S~\cite{wu2016tensorpack} on the SVHN dataset and the results of ConvNet~\cite{wu2016tensorpack} and VGG-nagadomi~\cite{vgg-nagadomi} on the CIFAR-10 dataset.
The accuracy of ConvNet-S on SVHN is slightly increased,
which is possibly due to that the low-precision computation decreases the overfitting of neural networks.
As can be seen, the accuracy loss of ConvNet, as well as VGG-nagadomi, on CIFAR-10 is trivial with 8-bit quantization.

\vspace{-0.2cm}
\begin{table}[!ht]
\centering
\scriptsize
\begin{tabular}{|c||c|c|c|c|c|c|c|}
\hline
\multicolumn{8}{|c|}{\textbf{ConvNet}} \\ \hline
\textbf{} & \textbf{BNN} & \textbf{BWN} & \textbf{TWN} & \multicolumn{3}{c|}{\textbf{LANCE (Ours)}} & \textbf{FULL} \\ \hline
\textbf{W-I} & 1-1 & 1-32 & 2-32 & 4-4 & 4-32 & 8-8 & 32-32 \\ \hline
\textbf{ACC} & 0.454 & 0.862 & 0.871 & 0.8669 & 0.8919 & \textbf{0.9016} & 0.8912 \\ \hline \hline
\multicolumn{8}{|c|}{\textbf{VGG-nagadomi}} \\ \hline
\textbf{} & \textbf{BNN} & \textbf{BWN} & \textbf{TWN} & \multicolumn{3}{c|}{\textbf{LANCE (Ours)}} & \textbf{FULL} \\ \hline
\textbf{W-I} & 1-1 & 1-32 & 2-32 & 4-4 & 4-32 & 8-8 & 32-32 \\ \hline
\textbf{ACC} & 0.734 & 0.874 & 0.887 & 0.8694 & 0.8908 & \textbf{0.9034} & 0.9042 \\ \hline
\end{tabular}
\caption{Accuracy comparison on CIFAR-10.}
\label{result-other}
\end{table}
\vspace{-0.2cm}
Table \ref{result-other} shows the accuracies under different quantization methods, including BNN~\cite{leng2018extremely}, BWN~\cite{hubara2017quantized}, and TWN~\cite{li2016ternary}.
As can be seen, our 8-8 quantization outperforms other methods, which even uses full-precision input and low-precision weights, and the computation of 8-bit data can be accelerated by low-precision computing on GPUs.
\vspace{-0.2cm}
\begin{figure}[!ht]
  \centering
  \includegraphics[width=0.5\textwidth]{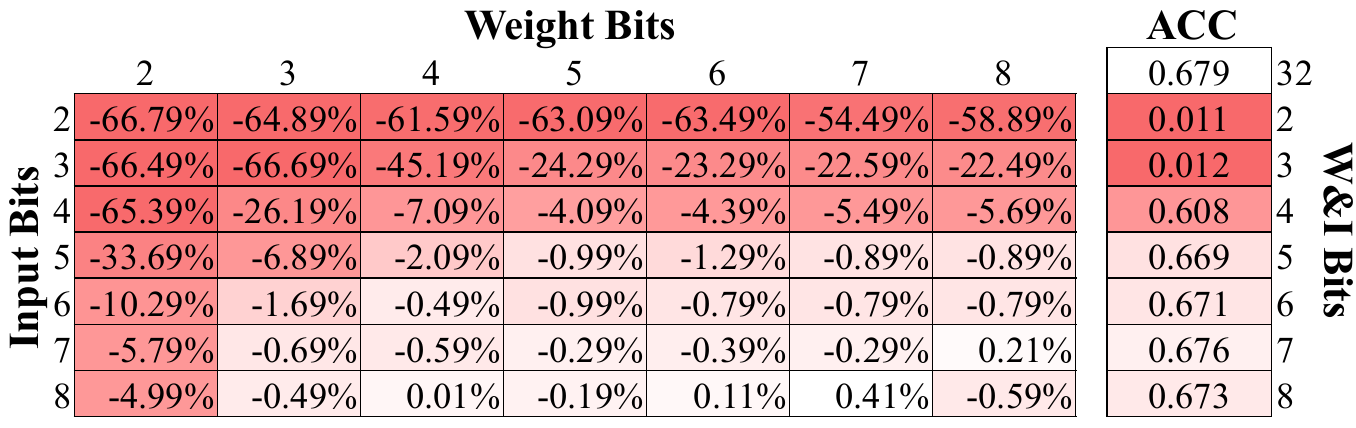}
  \caption{The result of ConvPool-CNN on CIFAR-100.}\label{result-cifar100}
\end{figure}
\vspace{-0.2cm}

Figure \ref{result-cifar100} illustrates the results of ConvPool-CNN~\cite{springenberg2014striving} on the CIFAR-100 dataset.
The experimental results are colored according to the accuracy loss.
In this experiment, we test the quantized Winograd convolution with different bit-widths of input and weights.
As illustrated, the accuracy loss of neural networks decreases with more bit-widths.

\vspace{-0.2cm}
\begin{table}[!h]
\centering
\scriptsize
\begin{tabular}{|c||c|c|c|c|c|}
\hline
\textbf{W-I}  & \textbf{8-8} & \textbf{7-7} & \textbf{6-6} & \textbf{5-5} & \textbf{4-4} \\ \hline
\textbf{TOP-1 ACC} & \textbf{$\uparrow$ 0.15\%} & $\uparrow$ 0.04\% & $\downarrow$ 0.21\% & $\downarrow$ 0.45\% & $\downarrow$ 3.00\% \\ \hline
\textbf{TOP-5 ACC} & \textbf{$\downarrow$ 0.04\%} & $\uparrow$ 0.03\% & $\downarrow$ 0.15\% & $\downarrow$ 0.55\% & $\downarrow$ 2.26\% \\ \hline
\end{tabular}
\caption{The result of ResNet-18 on ImageNet.}
\label{result-imagenet}
\end{table}
\vspace{-0.2cm}
We also test a variation of ResNet-18~\cite{he2016deep,DBLP:conf/iclr/LiuPHD18} on the ImageNet, which is a very large dataset.
As shown in Table \ref{result-imagenet}, the top-1 accuracy varies less than 0.2\% and the top-5 accuracy varies less than 0.1\% with 8-8 quantization.

\vspace{-0.2cm}
\begin{figure}[!ht]
  \centering
  \includegraphics[width=0.49\textwidth]{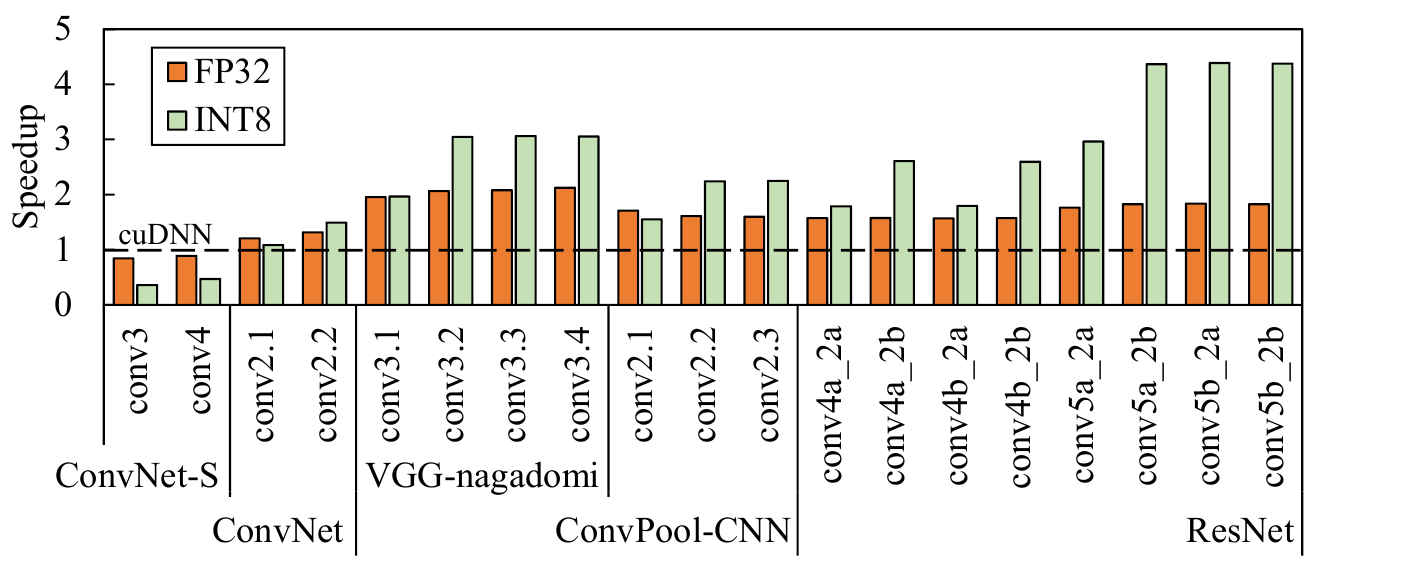}
  \caption{The speedup of main convolution layers.}
  \label{result-speedup}
\end{figure}
\vspace{-0.2cm}

Figure \ref{result-speedup} depicts the speedup of our method.
Our 8-bit quantized Winograd convolution improves the performance by up to 2.40$\times$ over the full-precision and 4.39$\times$ over the cuDNN implicit GEMM convolution.
The speedup increases with more filters and larger input size.
In general, the time of linear quantization operations is further less than the product computation.
We note that the performance of some layers is not improved, which is due to their very small sizes and the overhead of quantization cannot be neglected.

\noindent\textbf{Disscussion.}
The experimental results confirm the efficiency of our \textsc{Lance}.
By using 8-8 linear quantization, the performance of neural networks is significantly improved with trivial accuracy loss on datasets with different scales.
Using non-linear quantization methods may improve the performance, which remains as our future work.

\vspace{-0.2cm}
\section{Related Work}
\vspace{-0.2cm}
The efficient implementations of the Winograd convolution have been designed on different devices, such as mobile and edge devices~\cite{xygkis2018efficient, maji2019efficient}.
The specific hardware for Winograd convolution has also been proposed \cite{yang2018reconfigurable,lu2018spwa, wang2019low}.
Moreover, several researches focus on increasing the sparsity of the Winograd convolution using neural network pruning methods \cite{DBLP:conf/iclr/LiuPHD18, yu2019spatial, choi2019jointly}, which is complementary to our work.

\vspace{-0.2cm}
\section{Conclusion}
\vspace{-0.2cm}

In this paper, we proposed \textsc{Lance}, an efficient quantized Winograd convolution algorithm on graphics processing units.
The experimental results show that the \textsc{Lance} fully exploits the potential of low-precision computation by embedding quantization techniques, achieving significant speedup with trivial accuracy loss.

\newpage
{
\small
\bibliographystyle{IEEEbib}
\bibliography{refs}
}
\end{document}